\documentclass{article}

\PassOptionsToPackage{numbers, compress}{natbib}

\usepackage[preprint]{neurips_2022}
\usepackage{framed}
\usepackage{dirtytalk}
\usepackage{soul}

\usepackage[utf8]{inputenc}
\usepackage[T1]{fontenc}
\usepackage{multirow}
\usepackage{url} 
\usepackage{booktabs}

\title{mRobust04: A Multilingual Version of the TREC Robust 2004 Benchmark}
\author{Vitor Jeronymo, Mauricio Nascimento, Roberto Lotufo and Rodrigo Nogueira\\
NeuralMind, Brazil\\
FEEC, UNICAMP, Brazil}

\begin{document}

\maketitle

\begin{abstract}
    Robust 2004 is an information retrieval benchmark whose large number of judgments per query make it a reliable evaluation dataset.
    In this paper, we present mRobust04, a multilingual version of Robust04 that was translated to 8 languages using Google Translate. We also provide results of three different multilingual retrievers on this dataset. The dataset is available at \url{https://huggingface.co/datasets/unicamp-dl/mrobust}
\end{abstract}

\section{Introduction}

A key ingredient in the development of information retrieval algorithms are reusable evaluation datasets~\cite{craswell2021trec,voorhees2022too,voorhees2022can}.
For English, there are a number of such datasets. For other languages, there are initiatives such TREC CLIR~\cite{schauble1998cross}, CLEF~\cite{peters2001european,braschler2001clef,braschler2002clef,braschler2003clef}, FIRE~\cite{mitra2008overview,majumder2013overview}, NTCIR~\cite{sakai2021evaluating} and more recently HC4~\cite{lawrie2022hc4}.
A common problem with these multilingual IR datasets is their low number of judgments per query, that is, the number of documents marked as relevant or not relevant per query. For example, in the multilingual datasets mMARCO~\cite{mmarco} and Mr.Tydi~\cite{zhang2021mr}, there is only one or two documents manually marked as relevant per query. These "sparse" annotations, we argue, prevent correct evaluations of retrieval methods. For example, the RM3 query expansion method evaluated on the MS MARCO benchmark~\cite{MS_MARCO_v3}, which uses sparse annotations, shows no improvement over baselines such as BM25~\cite{lin2021pretrained}. However, the same method shows significant improvements over BM25 when evaluated on densely annotated benchmarks, such as TREC-DL.

In this work, to mitigate the issue with sparse annotations on multilingual IR datasets, we translate the TREC's Robust 2004 benchmark~\cite{Voorhees_TREC2004_robust}, an English dataset with a high number of judgments, to 8 languages using a high-quality automatic translator. We call this dataset mRobust04.

In Table~\ref{tab:dataset_summary} we compare mRobust04 with other multilingual IR datasets. Despite having a modest number of queries and documents, mRobust04 has much more annotations per query, which, we conjecture, makes it a reliable benchmark for evaluating future multilingual retrieval models.  We also evaluate on this dataset two multilingual models that are close to the state of the art.


\section{Translation Methodology}

Robust04 is an English news dataset whose documents are in a single file, divided by <DOC> tags. Regular expressions were used to delimit and extract relevant text from each document within the file, and BeautifulSoup was used to clean any HTML tags that may have remained.

After that, we fed Robust04's queries and corpus to the Google Translate API to translate to the following languages: Chinese, French, German, Indonesian, Italian, Portuguese, Russian and Spanish. Some documents had more characters than the maximum allowed by the API, so we used the nltk's~\cite{bird2009natural} sentence tokenizer to split them into chunks of acceptable sizes and send them independently to the API. The resulting translations of these chunks were then concatenated to form the translated document. The annotated query-document pairs are the same for all languages.

\begin{table}
\centering
\begin{tabular}{lcrrrr} 
\toprule
\textbf{Dataset} & \textbf{MT} & \textbf{Langs} & \textbf{Queries} & \textbf{Docs} & \textbf{J/q}\\
\midrule
Mr. Tydi & No & 11 & 2k-16k & 136k-32M & 1.03 \\
mMARCO & Yes & 14 & 540k & 8.8M & 1.06 \\
CLEF 2001-2003 & No & 5 & 50-60 & 87k-454k & 27.93 \\
HC4 & No & 3 & 54-60 & 486k-4.7M & 54.36 \\
TREC-8 CLIR & No & 4 & 28 & 62k-242k & 206.75 \\
\midrule
mRobust04 (ours) & Yes & 9 & 249 & 528k & 1250.60\\
\bottomrule
\end{tabular}
\vspace{0.2cm}
\caption{Comparison of mRobust04 with other multilingual IR datasets. ``MT'' refers to whether the dataset was machine translated or not. J/q is the average number of judgments per query per language.}
\label{tab:dataset_summary}
\end{table}
\section{Evaluation}

We evaluate a sparse model (BM25), a dense model (mColBERT) and a reranker (mT5) on mRobust04. We use mT5 and mColBERT finetuned on mMARCO as provided by Bonifacio et al.~\cite{mmarco} and evaluate them on mRobust04 in a zero-shot manner.

Inference in each language was performed by creating windows of sentences, with Spacy\cite{spacy2}, of maximum 10 and stride of 5 for both mT5 and mColBERT, since their maximum tokenized length is lower than most documents in the corpus.
The sparse first-stage retrieval, BM25, is not limited by a maximum length; therefore, the corpus was indexed without any windowing.

We report nDCG@20 and R@1000. In preliminary experiments, we observed that mT5 and mColBERT have 6\% and 20\% fewer judged documents on average in their top 20 compared to BM25. Therefore we also report nDCG' (i.e., nDCG ``prime'')~\cite{sakai2007alternatives}, which does not penalize the model for retrieving unjudged query-document pairs.

\begin{table}
    \centering
\begin{tabular}{lrrrrrrrrrr} 
\toprule
\multicolumn{1}{l}{} & \textit{en} & \textit{fr} & \textit{pt}  & \textit{it} & \textit{id} & \textit{ru} & \textit{es} & \textit{de} & \textit{zh} & \textbf{avg}  \\
\midrule

\multicolumn{11}{c}{\textbf{nDCG@20}}\\

BM25     & 0.389                        & 0.389                        & 0.389                        & 0.387                        & 0.383                        & 0.372                        & 0.364                        & 0.333                        & 0.289                        & 0.367                         \\
mT5      & 0.466                        & 0.376                        & 0.391                        & 0.384                        & 0.374                        & 0.372                        & 0.402                        & 0.375                        & 0.358                        & 0.389                         \\
mColBERT & 0.362                        & 0.302                        & 0.323                        & 0.305                        & 0.287                        & 0.265                        & 0.309                        & 0.280                         & 0.262                        & 0.300                           \\ \midrule
\multicolumn{11}{c}{\textbf{nDCG'@20}}                                                                                                                                                                                                                                                                                                   \\
BM25     & 0.394                        & 0.418                        & 0.409                        & 0.411                        & 0.407                        & 0.403                        & 0.394                        & 0.372                        & 0.349                        & 0.396                         \\
mT5      & 0.486                        & 0.429                        & 0.439                        & 0.436                        & 0.432                        & 0.431                        & 0.454                        & 0.435                        & 0.418                        & 0.440                          \\
mColBERT & 0.414                        & 0.383                        & 0.401                        & 0.379                        & 0.367                        & 0.348                        & 0.389                        & 0.361                        & 0.345                        & 0.377                         \\ \midrule
\multicolumn{11}{c}{\textbf{R@1000}}                                                                                                                                                                                                                                    \\
BM25     & 0.649                        & 0.655                        & 0.657                        & 0.628                        & 0.649                        & 0.627                        & 0.640                        & 0.514                        & 0.517                        & 0.616                         \\
mColBERT & 0.597                        & 0.526                        & 0.549                        & 0.525                        & 0.510                         & 0.475                        & 0.547                        & 0.503                        & 0.423                        & 0.518     \\
\bottomrule
\end{tabular}
\vspace{0.1cm}
\caption{Main results in the mRobust04 dataset. \textsc{mT5} and  \textsc{mColBERT} were finetuned on mMARCO.}
\label{tab:mrobust}
\end{table}

The results are shown in Table~\ref{tab:mrobust}. Except for the Chinese language, the nDCG@20 results for BM25 in each language are very close to English, indicating that our automatic translation was able to retain the information present in the original documents successfully.
The average nDCG@20 of the mT5 reranker slightly surpasses BM25's. Looking at nDCG'@20, mT5 shows clear improvements over BM25. This suggests that there is a non-negligible amount of documents in top 20 reranked by mT5 that are relevant but were not annotated.

We expected the dense model, mColBERT, to be worse than the mT5 model and better than BM25 as observed by Bonifacio et al.~\cite{mmarco}. This is because mColBERT would be able to overcome the lexical matching problem that BM25 suffers from, as dense models could potentially represent semantically similar words closer to each other in the indexed embedding space. However, mColBERT did not perform well on mRobust04, staying behind both models for all metrics and languages. One possibility is that mColBERT is ``overfitted'' to mMARCO and was unable to generalize to a new domain. Another explanation is similar to that proposed by Rosa et al.~\cite{rosa2022no}: dense retrievers show great in-domain effectiveness but poor out-of-domain generalization. However, rigorously testing this hypothesis for the multilingual scenario is beyond the scope of this work.

\section*{Acknowledgments}
This research was partially funded by grants 2020/09753-5 and 2022/01640-2 from Fundação de Amparo à Pesquisa do Estado de São Paulo (FAPESP).
We also thank Google Cloud for credits to support this work.

\bibliographystyle{abbrvnat}
\bibliography{ref} 

\begin{thebibliography}{21}
\providecommand{\natexlab}[1]{#1}
\providecommand{\url}[1]{\texttt{#1}}
\expandafter\ifx\csname urlstyle\endcsname\relax
  \providecommand{\doi}[1]{doi: #1}\else
  \providecommand{\doi}{doi: \begingroup \urlstyle{rm}\Url}\fi

\bibitem[Bajaj et~al.(2018)Bajaj, Campos, Craswell, Deng, Gao, Liu, Majumder,
  McNamara, Mitra, Nguyen, Rosenberg, Song, Stoica, Tiwary, and
  Wang]{MS_MARCO_v3}
P.~Bajaj, D.~Campos, N.~Craswell, L.~Deng, J.~Gao, X.~Liu, R.~Majumder,
  A.~McNamara, B.~Mitra, T.~Nguyen, M.~Rosenberg, X.~Song, A.~Stoica,
  S.~Tiwary, and T.~Wang.
\newblock {MS} {MARCO}: {A Human Generated MAchine Reading COmprehension
  Dataset}.
\newblock \emph{arXiv:1611.09268v3}, 2018.

\bibitem[Bird et~al.(2009)Bird, Klein, and Loper]{bird2009natural}
S.~Bird, E.~Klein, and E.~Loper.
\newblock \emph{Natural language processing with Python: analyzing text with
  the natural language toolkit}.
\newblock " O'Reilly Media, Inc.", 2009.

\bibitem[Bonifacio et~al.(2021)Bonifacio, Jeronymo, Abonizio, Campiotti,
  de~Alencar~Lotufo, and Nogueira]{mmarco}
L.~H. Bonifacio, V.~Jeronymo, H.~Q. Abonizio, I.~Campiotti,
  R.~de~Alencar~Lotufo, and R.~Nogueira.
\newblock mmarco: {A} multilingual version of {MS} {MARCO} passage ranking
  dataset.
\newblock \emph{CoRR}, abs/2108.13897, 2021.
\newblock URL \url{https://arxiv.org/abs/2108.13897}.

\bibitem[Braschler(2002{\natexlab{a}})]{braschler2001clef}
M.~Braschler.
\newblock Clef 2001 --- overview of results.
\newblock In C.~Peters, M.~Braschler, J.~Gonzalo, and M.~Kluck, editors,
  \emph{Evaluation of Cross-Language Information Retrieval Systems}, pages
  9--26, Berlin, Heidelberg, 2002{\natexlab{a}}. Springer Berlin Heidelberg.

\bibitem[Braschler(2002{\natexlab{b}})]{braschler2002clef}
M.~Braschler.
\newblock Clef 2002—overview of results.
\newblock In \emph{Workshop of the Cross-Language Evaluation Forum for European
  Languages}, pages 9--27. Springer, 2002{\natexlab{b}}.

\bibitem[Braschler(2003)]{braschler2003clef}
M.~Braschler.
\newblock Clef 2003--overview of results.
\newblock In \emph{Workshop of the cross-language evaluation forum for european
  languages}, pages 44--63. Springer, 2003.

\bibitem[Craswell et~al.(2021)Craswell, Mitra, Yilmaz, Campos, Voorhees, and
  Soboroff]{craswell2021trec}
N.~Craswell, B.~Mitra, E.~Yilmaz, D.~Campos, E.~M. Voorhees, and I.~Soboroff.
\newblock Trec deep learning track: reusable test collections in the large data
  regime.
\newblock In \emph{Proceedings of the 44th International ACM SIGIR Conference
  on Research and Development in Information Retrieval}, pages 2369--2375,
  2021.

\bibitem[Honnibal and Montani(2017)]{spacy2}
M.~Honnibal and I.~Montani.
\newblock {spaCy 2}: Natural language understanding with {B}loom embeddings,
  convolutional neural networks and incremental parsing.
\newblock To appear, 2017.

\bibitem[Lawrie et~al.(2022)Lawrie, Mayfield, Oard, and Yang]{lawrie2022hc4}
D.~Lawrie, J.~Mayfield, D.~W. Oard, and E.~Yang.
\newblock Hc4: a new suite of test collections for ad hoc clir.
\newblock In \emph{European Conference on Information Retrieval}, pages
  351--366. Springer, 2022.

\bibitem[Lin et~al.(2021)Lin, Nogueira, and Yates]{lin2021pretrained}
J.~Lin, R.~Nogueira, and A.~Yates.
\newblock Pretrained transformers for text ranking: Bert and beyond.
\newblock \emph{Synthesis Lectures on Human Language Technologies}, 14\penalty0
  (4):\penalty0 1--325, 2021.

\bibitem[Majumder et~al.(2013)Majumder, Pal, Bandyopadhyay, and
  Mitra]{majumder2013overview}
P.~Majumder, D.~Pal, A.~Bandyopadhyay, and M.~Mitra.
\newblock Overview of fire 2010.
\newblock In \emph{Multilingual Information Access in South Asian Languages},
  pages 252--257. Springer, 2013.

\bibitem[Mitra(2008)]{mitra2008overview}
M.~Mitra.
\newblock Overview of fire 2008.
\newblock In \emph{Working Notes of Forum for Information Retrieval
  Evaluation}, 2008.

\bibitem[Peters and Braschler(2001)]{peters2001european}
C.~Peters and M.~Braschler.
\newblock European research letter: Cross-language system evaluation: The clef
  campaigns.
\newblock \emph{Journal of the American Society for Information Science and
  Technology}, 52\penalty0 (12):\penalty0 1067--1072, 2001.

\bibitem[Rosa et~al.(2022)Rosa, Bonifacio, Jeronymo, Abonizio, Fadaee, Lotufo,
  and Nogueira]{rosa2022no}
G.~M. Rosa, L.~Bonifacio, V.~Jeronymo, H.~Abonizio, M.~Fadaee, R.~Lotufo, and
  R.~Nogueira.
\newblock No parameter left behind: How distillation and model size affect
  zero-shot retrieval.
\newblock \emph{arXiv preprint arXiv:2206.02873}, 2022.

\bibitem[Sakai(2007)]{sakai2007alternatives}
T.~Sakai.
\newblock Alternatives to bpref.
\newblock In \emph{Proceedings of the 30th annual international ACM SIGIR
  conference on Research and development in information retrieval}, pages
  71--78, 2007.

\bibitem[Sakai et~al.(2021)Sakai, Oard, and Kando]{sakai2021evaluating}
T.~Sakai, D.~W. Oard, and N.~Kando.
\newblock \emph{Evaluating Information Retrieval and Access Tasks: NTCIR's
  Legacy of Research Impact}.
\newblock Springer Nature, 2021.

\bibitem[Sch{\"a}uble and Sheridan(1998)]{schauble1998cross}
P.~Sch{\"a}uble and P.~Sheridan.
\newblock Cross-language information retrieval (clir) track overview.
\newblock \emph{NIST SPECIAL PUBLICATION SP}, pages 31--44, 1998.

\bibitem[Voorhees(2004)]{Voorhees_TREC2004_robust}
E.~M. Voorhees.
\newblock Overview of the {TREC} 2004 robust track.
\newblock In \emph{Proceedings of the Thirteenth Text REtrieval Conference
  (TREC 2004)}, pages 52--69, Gaithersburg, Maryland, 2004.

\bibitem[Voorhees et~al.(2022{\natexlab{a}})Voorhees, Craswell, and
  Lin]{voorhees2022too}
E.~M. Voorhees, N.~Craswell, and J.~Lin.
\newblock Too many relevants: Whither cranfield test collections?
\newblock In \emph{Proceedings of the 45th Annual International ACM SIGIR
  Conference on Research and Development in Information Retrieval (SIGIR
  2022)}, 2022{\natexlab{a}}.

\bibitem[Voorhees et~al.(2022{\natexlab{b}})Voorhees, Soboroff, and
  Lin]{voorhees2022can}
E.~M. Voorhees, I.~Soboroff, and J.~Lin.
\newblock Can old trec collections reliably evaluate modern neural retrieval
  models?
\newblock \emph{arXiv preprint arXiv:2201.11086}, 2022{\natexlab{b}}.

\bibitem[Zhang et~al.(2021)Zhang, Ma, Shi, and Lin]{zhang2021mr}
X.~Zhang, X.~Ma, P.~Shi, and J.~Lin.
\newblock Mr. tydi: A multi-lingual benchmark for dense retrieval.
\newblock In \emph{Proceedings of the 1st Workshop on Multilingual
  Representation Learning}, pages 127--137, 2021.

\end{thebibliography}

\end{document}